\documentclass[conference]{IEEEtran}
\IEEEoverridecommandlockouts
\usepackage{cite}
\usepackage{amsmath,amssymb,amsfonts}
\usepackage{algorithmic}
\usepackage{graphicx}
\usepackage{textcomp}
\usepackage{xcolor}
\def\BibTeX{{\rm B\kern-.05em{\sc i\kern-.025em b}\kern-.08em
    T\kern-.1667em\lower.7ex\hbox{E}\kern-.125emX}}
\begin{document}

\title{Quantized Context Based LIF Neurons for Recurrent Spiking Neural Networks in 45nm
\thanks{*Equal Contribution}
\thanks{Accepted at NICE 2024 (Neuro-Inspired Computational Elements) Conference 2024, Apr. 23-26, 2024 }
}

\author{
    \IEEEauthorblockN{Sai Sukruth Bezugam\IEEEauthorrefmark{1}\IEEEauthorrefmark{2}, Yihao Wu\IEEEauthorrefmark{1}, JaeBum Yoo\IEEEauthorrefmark{1}, Dmitri Strukov\IEEEauthorrefmark{3}, Bongjin Kim\IEEEauthorrefmark{4}}
    \IEEEauthorblockA{\textit{Department of Electrical and Computer Engineering} \\
    \textit{University of California, Santa Barbara}\\
    California, USA \\
    \IEEEauthorrefmark{2}saisukruthbezugam@ieee.org,
    \IEEEauthorrefmark{3}strukov@ece.ucsb.edu, 
    \IEEEauthorrefmark{4}bongjin@ucsb.edu}
}

\maketitle

\begin{abstract}
In this study, we propose the first hardware implementation of a context-based recurrent spiking neural network (RSNN) emphasizing on integrating dual information streams within the neocortical pyramidal neurons specifically Context-Dependent Leaky Integrate and Fire (CLIF) neuron models, essential element in RSNN. We present a quantized version of the CLIF neuron (qCLIF), developed through a hardware-software codesign approach utilizing the sparse activity of RSNN. Implemented in a 45nm technology node, the qCLIF is compact (900um²) and achieves a high accuracy of 90\% despite 8 bit quantization on DVS gesture classification dataset. Our analysis spans a network configuration from 10 to 200 qCLIF neurons, supporting up to 82k synapses within a 1.86 mm² footprint, demonstrating scalability and efficiency.  
\end{abstract}


\section{Introduction}
As the demand for more efficient and capable computing systems grows, neuromorphic computing has emerged as a promising avenue for emulating brain-like processing capabilities. This field, bridging artificial intelligence and neuroscience, not only aims to replicate human brain functions but also seeks to drastically reduce the power consumption of computational systems \cite{christensen20222022}, \cite{schuman2022opportunities}. In this paper, we explore how the integration of advanced neuron models can potentially address these challenges. Spiking Neural Networks (SNNs), termed the third generation of neural networks, offer a pathway to this goal through spike-based computations. However, many SNNs have not yet achieved the accuracy levels of ANNs. Efforts to bridge the performance gap have explored various approaches, including exploiting the multi-timescale dynamics of neurons \cite{Bellec2018, Shaban2021,Ganguly2024}, local learning algorithms \cite{Bellec2020, Yin2023}. A notable advancement in this domain is the integration of context-dependent leaky integrate and fire (CLIF) neurons into recurrent spiking neural networks (RSNNs) \cite{ferrand2023context, Baronig2023}. 
\begin{figure}
    \centering
    \includegraphics[width=0.9\linewidth]{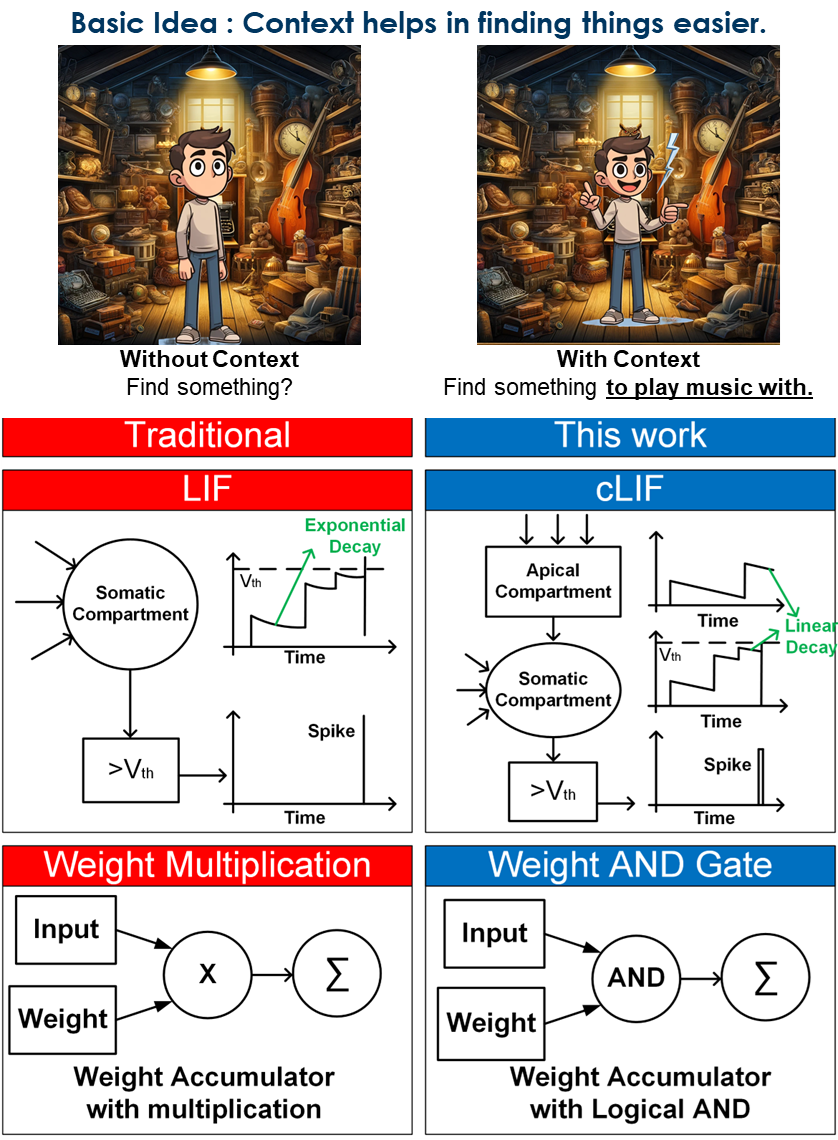}
    \caption{Top: Basic idea of using context.  Bottom : Comparison of Neuronal Models and Synaptic Processing: Traditional LIF versus proposed qCLIF. The Traditional LIF model exhibits a single-compartment with exponential decay and spike generation upon reaching threshold voltage, with an analog weight multiplication approach. The qCLIF model introduces an additional apical compartment with linear decay dynamics and utilizes a digital weight AND gate mechanism for synaptic processing, offering advantages in speed, reconfigurability, robustness, and scalability of digital hardware methodology.}
    \label{fig:over}
\end{figure}

The fundamental concept posits that understanding or locating items becomes more manageable with appropriate context. Consider the scenario of entering a disorganized room and being asked to find an object; the task proves challenging, overwhelmed by numerous options, making decision-making difficult. However, if the request specifies context, such as "find something to play music with," the search simplifies due to the targeted nature of the inquiry. This process involves two distinct streams of information: 1) stimuli information and 2) context. Though independent, correlating these streams enhances object identification efficiency. Drawing parallels with neocortical pyramidal neurons, this method employs dual information pathways: the bottom-up stimuli received by basal dendrites and the top-down context provided to apical tufts. Each pathway processes separate information—basal dendrites handle stimuli while apical tufts manage context. A correlation between these streams may trigger an higher output frequency. This integration has improved accuracy in gesture classification and speech recognition tasks, outperforming several existing models \cite{slayer,5,6,7,8,9}.

\begin{figure}
    \centering
    \includegraphics[width=\linewidth]{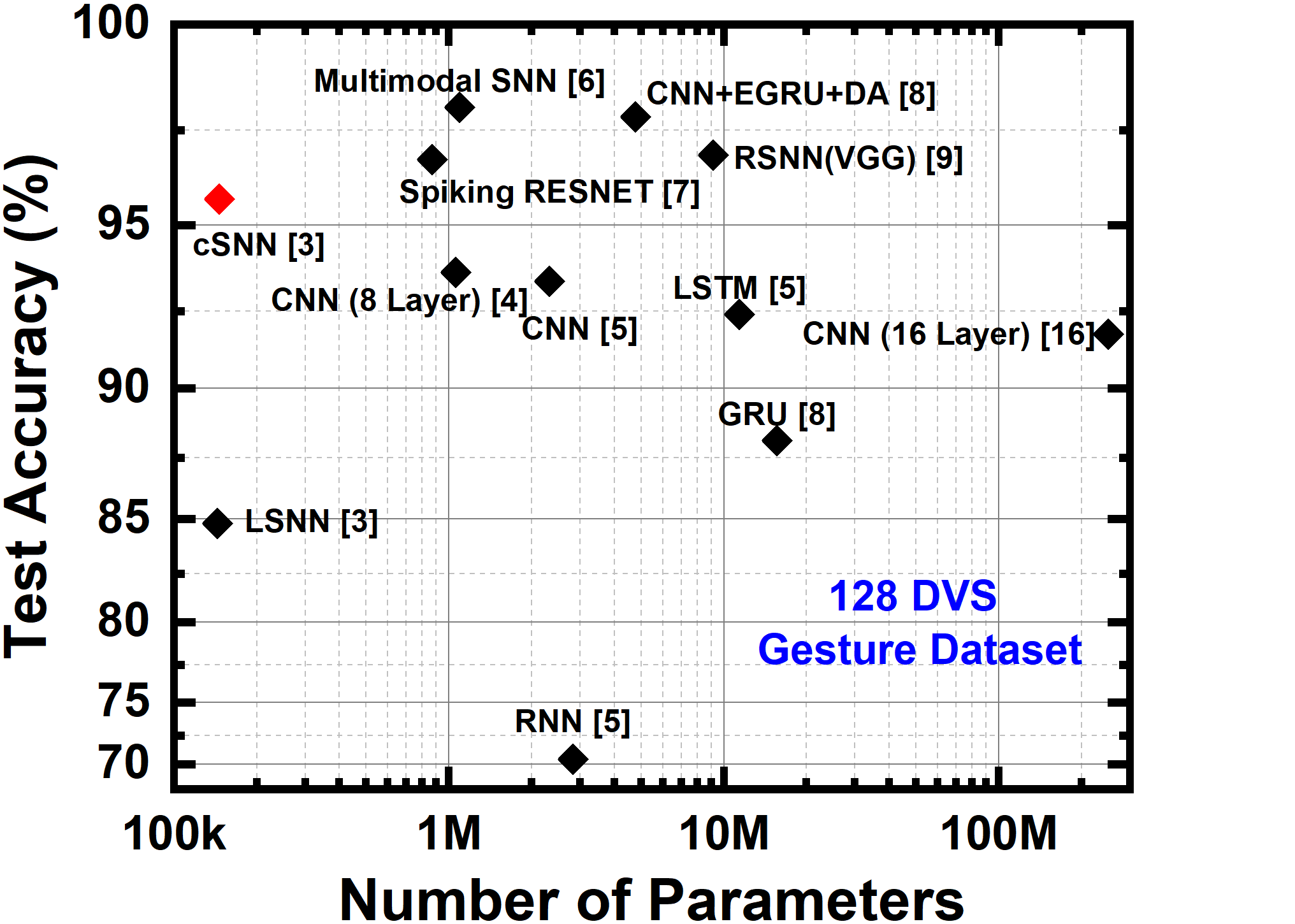}
    \caption{A comparison of test accuracy versus parameter count for diverse network architectures addressing the DVS Gesture dataset. Notably, context based Recurrent Spiking Neural Networks (cSNNs) achieve high accuracy with significantly fewer parameters compared to other models referenced in the literature.}
    \label{fig:fig2a}
\end{figure}

The unique feature of CLIF neurons in RSNNs is their use of contextual input to enhance the somatic compartment's computational capacity (see Fig. \ref{fig:over}). Using this additional stream of information as context these networks give accuracy on par with most of the network which are much larger in size Fig. \ref{fig:fig2a}. There have been no hardware implementations of the CLIF neuron model, although analog and digital versions of other neuron models exist. This work introduces a hardware-friendly variant of the CLIF neuron: the Quantized CLIF neuron. This model retains the accuracy of the original while being more amenable to digital implementation. The study systematically analyzes the model's performance with quantized neuron parameters and weights and examines network activity patterns.

\begin{figure*}[!htb]
    \centering
    \includegraphics[width=\linewidth]{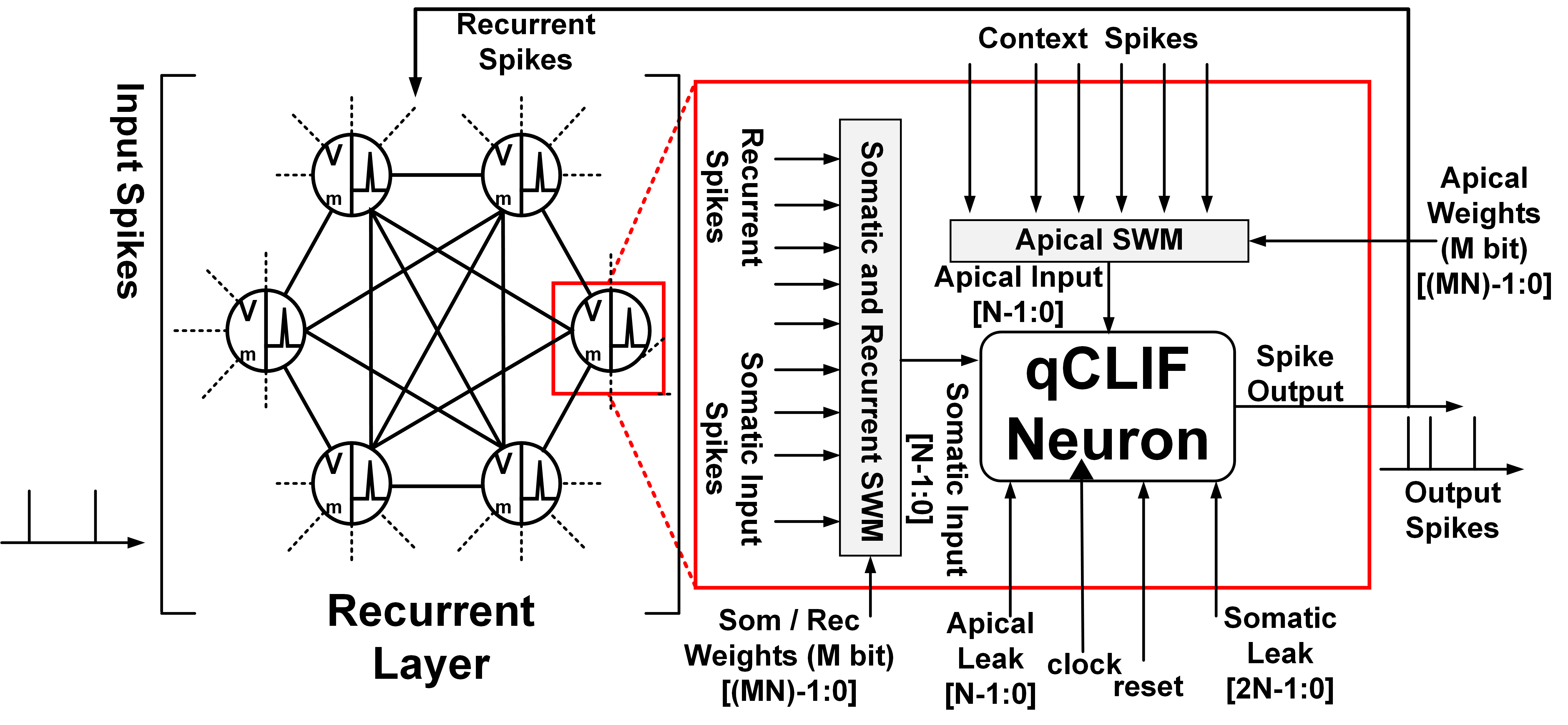}
    \caption{Architecture of a Recurrent Spiking Neural Network layer made of qCLIF Neurons. The inset view highlights a single qCLIF neuron, which processes both input and recurrent spikes using a combination of somatic and apical inputs along with somatic and recurrent, apical weights. It operates with a set of parameters including somatic leak and apical leak, governed by a clock and reset mechanism generating a train of output spikes.}
    \label{fig:arch}
\end{figure*}

The proposed neuron model, including its synaptic compartment, was implemented using open source 45 nm technology \cite{10}. Layout and post-synthesis evaluations assessed area, power, and timing characteristics. These assessments provide critical insights into the feasibility and efficiency of implementing the qCLIF neuron model in neuromorphic hardware.  The following sections will elaborate on the detailed methodology, results, comprehensive qCLIF neuron model analysis, hardware synthesis, and performance evaluations.

\section{proposed quantized context-based leaky integrate and fire neuron model}

\subsection{Mathematical Model}

Neuronal computation is segregated into two primary domains: the apical compartment, which assimilates contextual information, and the somatic compartment, which processes the primary stimulus inputs like spikes from various sensory modalities. The classical CLIF neuron model \cite{ferrand2023context} captures this bifurcation with equations that reflect the dynamic interplay between these compartments, as shown in eq. (\ref{eq:eq1}) (\ref{eq:eq2}).

\begin{equation}
V_j^a (t+\Delta t) = \alpha V_j^a (t) + (1-\alpha) R_m I_j^a (t+\Delta t)
\label{eq:eq1}
\end{equation}

\begin{equation}
\begin{split}
V_j (t+\Delta t) = & \beta V_j (t) + (1-\beta) [R_m I_j (t+\Delta t) \cdot \\
& \text{ReLU}(V_j^a (t+\Delta t))] - V_{th}
\end{split}
\label{eq:eq2}
\end{equation}

where $\alpha = \exp\left(-\frac{\Delta t}{\tau_a}\right)$ and $\beta = \exp\left(-\frac{\Delta t}{\tau_m}\right)$ are the exponential decay constants for the apical and somatic potentials, respectively. $\Delta t$ is typically set at 1 ms, akin to biological neurons. $I_j^a$ and $I_j$ represent apical and somatic (stimuli) input currents. $R_m$ denotes the membrane resistance, $V_{th}$ is the spiking threshold, and $s_j(t)$, which can be either 1 (indicating a spike) or 0 (no spike) $V_j$  greater than or less than $V_{th}$ respectively. To tailor these dynamics for digital systems, we adjust $\Delta t$ to equate to a single simulation timestep or clock cycle, which aligns the model with the discrete nature of digital computation. Further, we optimize by approximating many computational steps. The proposed qCLIF neuron model is expressed in eq. (\ref{eq:eq3}) (\ref{eq:eq4}).

\begin{equation}
V^{ap}(t + dt) = V^{ap}(t) - \alpha_{\text{leak}} + V_{\text{input}}^{ap}(t + dt)
\label{eq:eq3}
\end{equation}

\begin{equation}
\begin{split}
V^{\text{som}}(t + dt) = & V^{\text{som}}(t) - \beta_{\text{leak}} \\
& + \left(\text{RELU}\left(V^{ap}(t + dt)\right) \cdot V_{\text{input}}^{\text{som}}(t + dt)\right)
\end{split}
\label{eq:eq4}
\end{equation}

\(\alpha_{\text{leak}}\) and \(\beta_{\text{leak}}\) are the linear decay constants. \(V_{\text{input}}^{\text{ap}}(t + dt)\) and \(V_{\text{input}}^{\text{som}}(t + dt)\) are the contextual and stimulus inputs chosen to be of ‘\textit{N}’ bit-width fixed-point numbers, respectively. A new issue arises with constant linear leakage: it can cause compartment voltages to fall below zero uncontrollably. To address this, we set the lower limit of the voltages to zero. Upon neuron spiking, a 'Reset to Zero' mechanism is applied to the somatic compartment. These inputs are defined by eq. (\ref{eq:eq5}) (\ref{eq:eq6}).

\begin{equation}
V_{\text{input}}^{\text{con}}(t + dt) = \sum (\text{AND}(C_{\text{spike}}, W_{\text{context}}))
\label{eq:eq5}
\end{equation}

\begin{equation}
\begin{split}
V_{\text{input}}^{\text{som}}(t + dt) = & \sum (\text{AND}(S_{\text{spike}}, W_{\text{soma}})) \\
& + \sum (\text{AND}(P_{\text{spike}}, W_{\text{recurrent}}))
\end{split}
\label{eq:eq6}
\end{equation}

\(C_{\text{spike}}\) and \(W_{\text{spike}}\) represent the contextual and somatic spike inputs, \(W_{\text{context}}\), \(W_{\text{soma}}\), and \(W_{\text{recurrent}}\) are the corresponding synaptic weights, and \(P_{\text{spike}}\) denotes the previous spike information for recurrent connections. Despite these modifications, the fundamental characteristics of neuronal activity are retained. The model leverages a piecewise-linear approach to mimic the neuron's response, especially in operational ranges where linear and exponential decay patterns are virtually indistinguishable. Moreover, the model effectively captures the core processes of neuronal dynamics: it integrates and decays inputs within the apical and somatic compartments and enhances the somatic potential through interaction with the modulated apical input.  The next section will discuss the digital implementation of the proposed qCLIF model.
 
\subsection{Architecture}

\begin{figure}
    \centering
    \includegraphics[width=\linewidth]{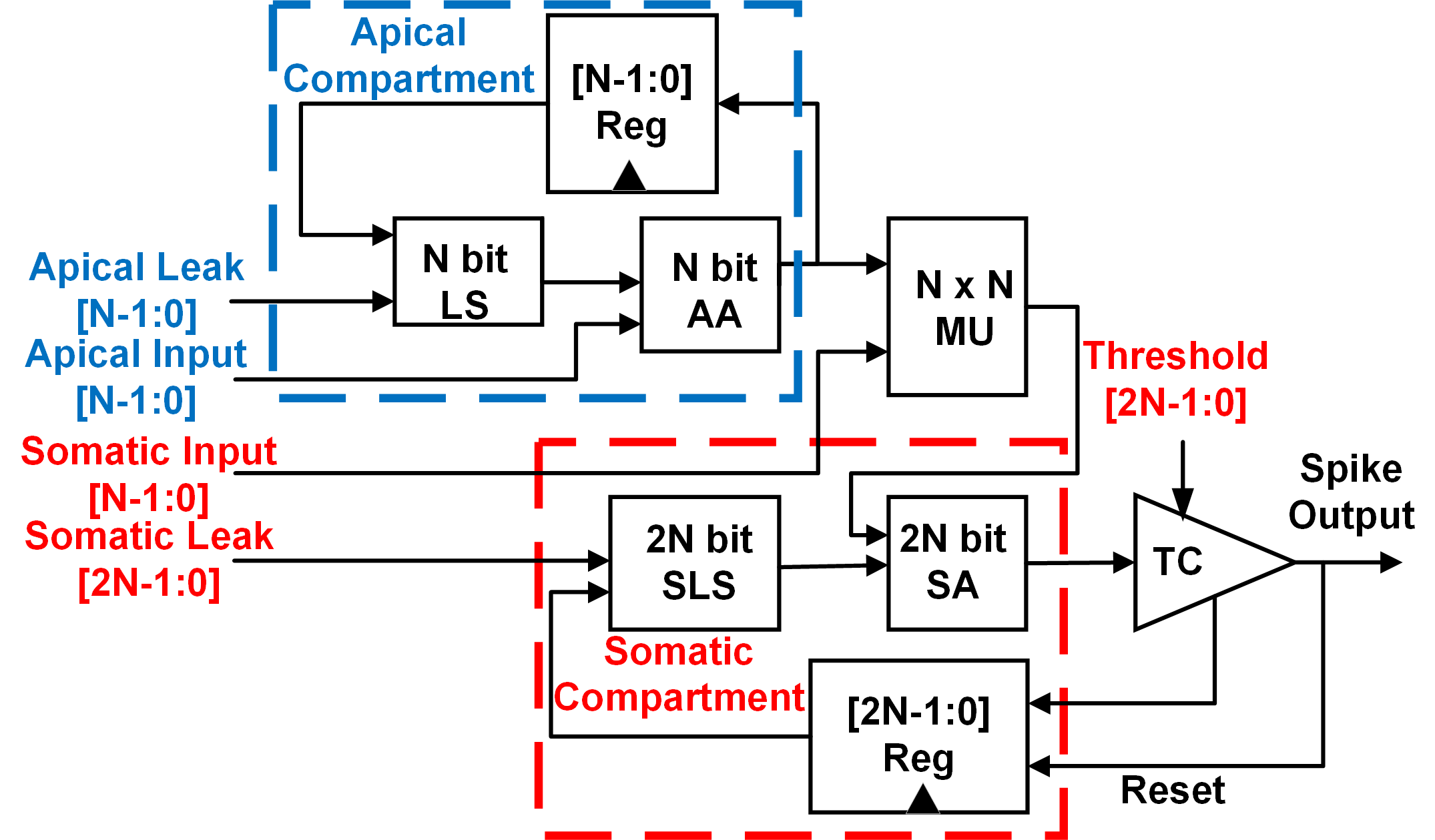}
    \caption{Digital design of \textit{N} bit qCLIF neuron. LS : Leakage Subtractor, AA: Apical Accumulator, MU: Multiplication Unit, SLS: Somatic Leakage Subtractor, SA: Somatic Accumulator, TC: Threshold Comparator.}
    \label{fig:qlif}
\end{figure}

For building a recurrent layer of qCLIF neurons, a modular approach is employed to process and integrate both external inputs and internally generated feedback, which is a key characteristic of the dynamics in recurrent neural network systems. This architecture ensures efficient spike processing and temporal data integration. The proposed architecture is outlined in Fig. \ref{fig:arch} and digital design of qCLIF is shown in Fig. \ref{fig:qlif}. The key modules in this architecture are detailed as follows:
\textbf{Spike Weighting Module (SWM):} The SWM processes incoming spikes from external somatic and apical stimuli. Additionally, it handles spikes generated internally by the network, which are used as inputs in the next step. Incorporating recurrent feedback is crucial for the temporal dynamics inherent in the network's processing. Each external or recurrent spike is combined with its corresponding \textit{M}-bit weight value using bitwise AND operations. The module employs Carry Save-Ahead (CSA) adders optimized for speed with a pipeline structured in $log_3(N)$ stages. \textbf{Apical Compartment (AC):} This component consists of a Leakage Subtractor (LS) and an Apical Accumulator (AA). The AC processes the outputs from the SWM, with the LS managing linear leakage from the accumulated data and the AA adding contextual inputs along with the recurrent feedback. Output registers connected to the AA store the cumulative sums, allowing for their reuse in subsequent accumulation cycles. The signed bit is checked in each clock cycle. Whenever the output is negative, the register is reset to $0$. \textbf{Multiplication Unit (MU):} The MU receives combined outputs from the AC, including external and internal data. It multiplies this data with additional stimuli inputs using an array multiplier, producing a $2N$-bit binary product that is then sent to the Somatic Compartment for further processing. \textbf{Somatic Compartment (SC):} Structurally similar to the AC, the SC comprises the Somatic Leakage Subtractor (SLS) and the Somatic Accumulator (SA). It accumulates the $2N$-bit data from the MU, with the SLS adjusting for leakage and the SA summing the inputs for threshold comparison. \textbf{Threshold Comparator (TC):} In the final stage, the TC compares the aggregate output from the SC against a predetermined threshold. If the output surpasses this threshold, a spike is generated. This output spike plays a dual role as both the neuron's output and an input for the SWM in the subsequent computational cycle, perpetuating the recurrent feedback loop within the network. Further, the spike is connected as \textit{RESET} to somatic compartment register.

\section{results}

\begin{figure}[h]
    \centering
    \includegraphics[width=0.8\linewidth]{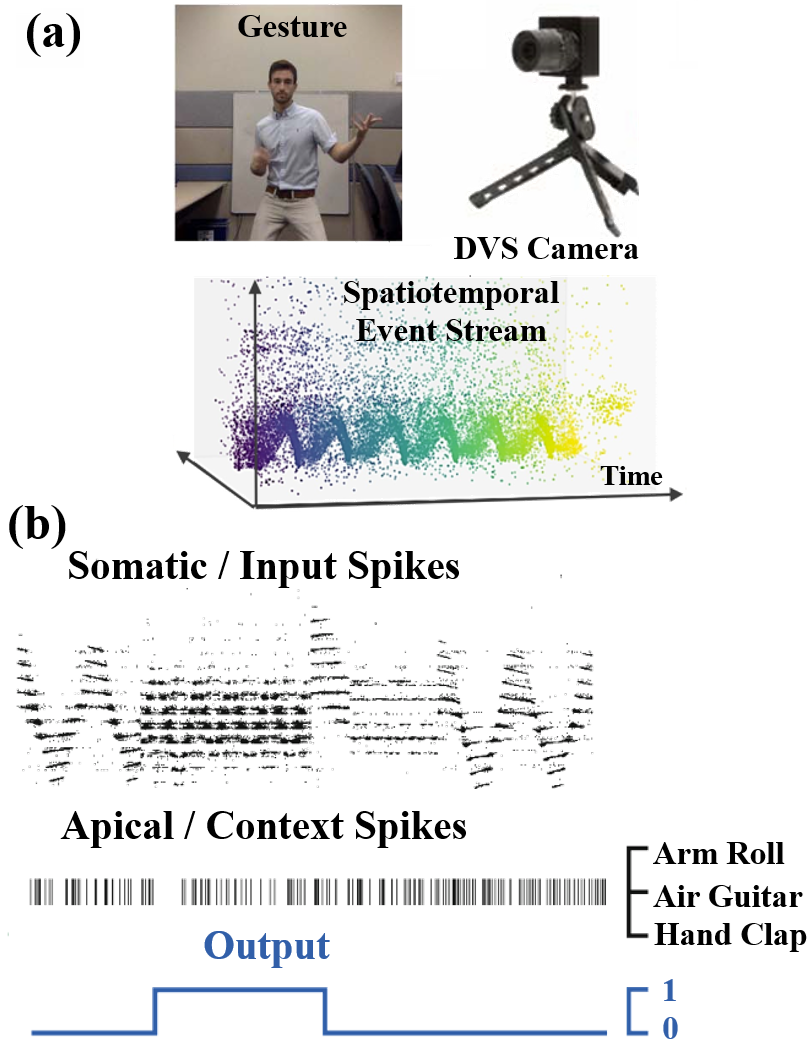}
    \caption{(a) Setup for Gesture Recognition Using DVS: This setup is utilized for recording gestures, which generate spatio-temporal event streams. The images are adapted from source \cite{11}. (b) Simulation Setup: This involves the use of a spatio-temporal stream as somatic input spikes. Context spikes are directed to the apical compartment, prompting the network to recognize a specific class. The output indicates whether the input stream corresponds with the context spikes.}
    \label{fig:dvsgesture}
\end{figure}

The efficacy of the proposed qCLIF neuron model was evaluated within a context-dependent RSNN using the Dynamic Vision Sensor (DVS) Gesture dataset \cite{11}. This dataset comprises ten distinct categories of hand gestures, each recorded with a spiking vision sensor. Inputs to the model are structured as 512-dimensional spike trains, with durations ranging from 196 to 1476 milliseconds. The network architecture consists of 200 cLIF neurons, connected recurrently through their somatic compartments. A context input mechanism incorporating ten neurons, each aligned with a specific class, was integrated. The target class is indicated by Poisson spikes at 200 Hz from the respective neuron, and the network's output is binary, indicating the correspondence of a gesture to the target class. The simulation setup The training was conducted over ten epochs using Backpropagation Through Time (BPTT) with an Adam optimizer. In Fig. \ref{fig:dvsgesture}, we present both the practical setup for gesture recognition using a DVS and the corresponding simulation framework, illustrating how spatio-temporal event streams are recorded and processed for gesture classification. For this task, we have determined that both somatic leak and apical leak are uniformly distributed across neurons, with values of (200, 7) and 7 respectively. This distribution is aimed at aligning their exponential decay parameters at 200 $\delta t$ and 20 $\delta t$. Additionally, we have adopted other parameters from \cite{ferrand2023context}.

\begin{figure}
    \centering
    \includegraphics[width=\linewidth]{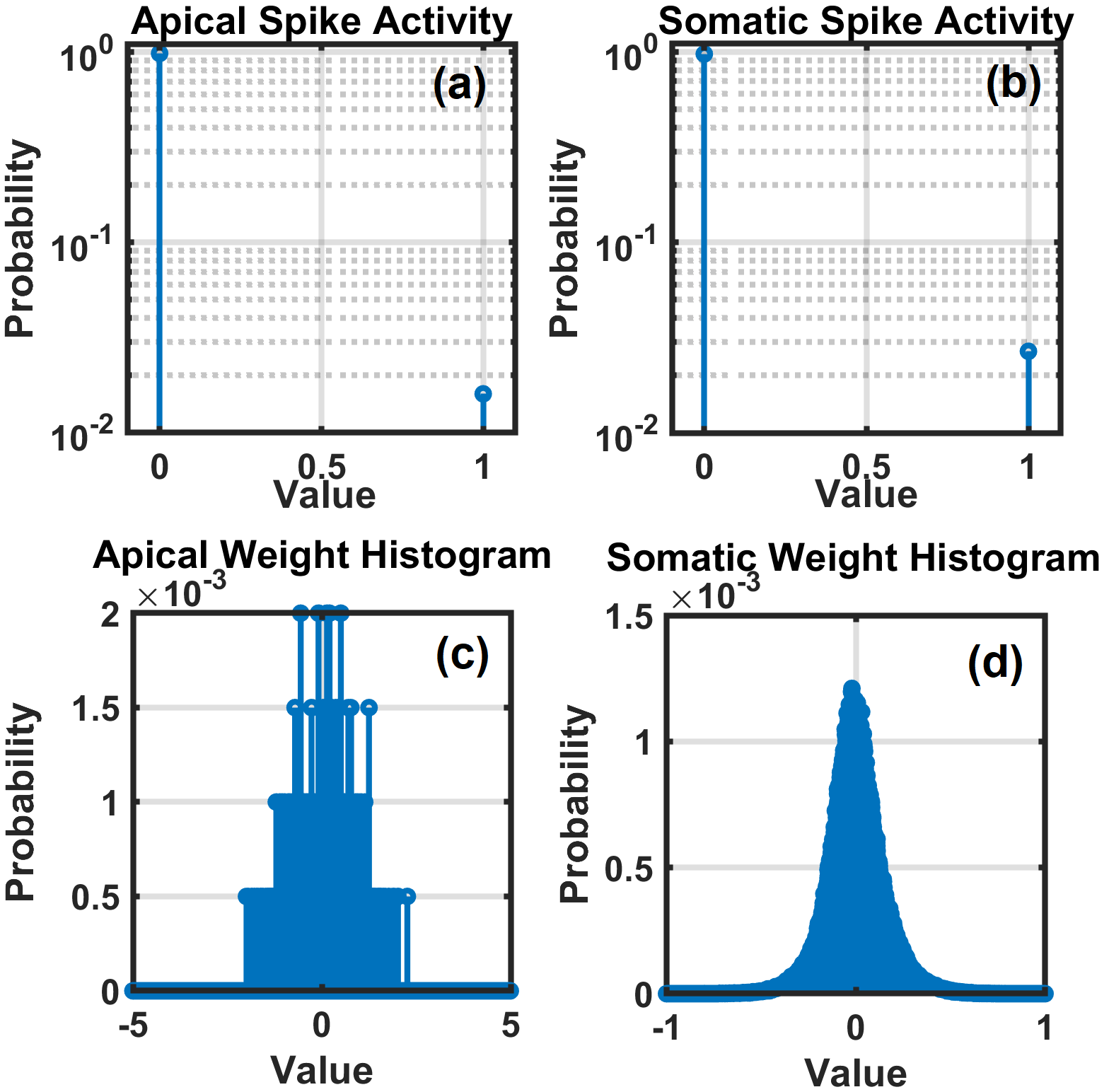}
    \caption{Insights from the network simulation (a) Apical input spike activity histogram, (b) Somatic input spike activity histogram, (c) Apical weight Histogram, (d) Somatic weight histogram.}
    \label{fig:quan}
\end{figure}

Initial tests focused on the impact of linear leakage of qCLIF on network performance. With optimized leakage parameters (choosing a linear regime of exponential decay), the qCLIF model's performance was found to be slightly below the state-of-the-art accuracy of 95\% by 0.5\%, proving its efficacy. Furthermore, removing leakage from the neuron model led to a notable 4\% decrease in accuracy at full precision.

The potential of quantization was also examined through fixed point numbers for constants and variables in eq. (\ref{eq:eq5}) (\ref{eq:eq6}). Considering over 500 inputs per neuron (context spikes, stimuli spikes, and recurrent spikes) processed through associated weights (assumed quantized to 8 bits), an adder precision of at least 16 bits was deemed necessary (to handle a maximum total of 127,500). This would necessitate a 16x16 multiplier and a final 32-bit adder. However, an in-depth analysis of network activity showed that spiking occurred in only about 2\% of the neuron population, as depicted in Fig. \ref{fig:quan} (a-b). This finding justified the use of smaller bit widths.

\begin{table}[]
\centering
\caption{EFFECT OF QUANTIZATION ON NETWORK PERFORMANCE}
\begin{tabular}{|c|c|c|}
\hline
\textbf{\begin{tabular}[c]{@{}c@{}}Precision\\ Level\end{tabular}} &
  \textbf{\begin{tabular}[c]{@{}c@{}}Neuron\\ Quantization\\ Accuracy (\%)\end{tabular}} &
  \textbf{\begin{tabular}[c]{@{}c@{}}Weight\\ and\\ Neuron\\ Quantization\\ Accuracy (\%)\end{tabular}} \\ \hline
\textbf{Full Precision} & 94.5 & 94.5 \\ \hline
\textbf{16-bit}         & 93.4 & 93   \\ \hline
\textbf{8-bit}          & 92   & 90   \\ \hline
\textbf{4-bit}          & 77.5 & 73   \\ \hline
\textbf{2-bit}          & 55   & N/A  \\ \hline
\end{tabular}
\label{tab:quan}
\end{table}

The effects of quantizing neurons and weights on network performance were also examined. The weight distribution within the network was observed to follow a normal distribution, with soma weights predominantly in the range of -0.5 to 0.5 and apical weights between -2 and 2. This informed the decision to quantize weights within these specific ranges, deviating from the common -1 to 1 range, as shown in Fig. \ref{fig:quan} (c-d). This tailored approach to quantization aimed to fully utilize the range, thus enhancing network efficiency and effectiveness. The consequent impact of quantization on model performance is elaborated in Table \ref{tab:quan}.

\begin{table*}[]
\centering
\caption{QCLIF NEURAL MODEL COMPUTING PERFORMANCE METRICS (@ 1.1V)}
\begin{tabular}{|c|c|c|c|c|c|c|c|c|c|}
\hline
\textbf{\begin{tabular}[c]{@{}c@{}}No. \\ of \\ qCLIF\end{tabular}} &
  \textbf{\begin{tabular}[c]{@{}c@{}}Clock\\ Freq\\ ($MHz$)\end{tabular}} &
  \textbf{\begin{tabular}[c]{@{}c@{}}Synapse\\ Count, Precision\end{tabular}} &
  \textbf{\begin{tabular}[c]{@{}c@{}}Area\\ ($mm^2$)\\ ($L X W$)\end{tabular}} &
  \textbf{\begin{tabular}[c]{@{}c@{}}Slack\\ (ns)\end{tabular}} &
  \textbf{\begin{tabular}[c]{@{}c@{}}Switching\\ Power \\ ($mW$)\end{tabular}} &
  \textbf{\begin{tabular}[c]{@{}c@{}}Internal\\ Power\\ ($mW$)\end{tabular}} &
  \textbf{\begin{tabular}[c]{@{}c@{}}Leakage\\ Power \\ ($mW$)\end{tabular}} &
  \textbf{\begin{tabular}[c]{@{}c@{}}Total \\ Power\\ ($mW$)\end{tabular}} &
  \textbf{\begin{tabular}[c]{@{}c@{}}Energy\\ Per \\ Spike\\ ($pJ$)\end{tabular}} \\ \hline
1   & 100                      & -         & 0.029 X 0.030 & 5.62  & 0.020   & 0.041 & 0.016 & 0.077 & 0.773 \\ \hline
10  & 20                       & 250, 8bit & 0.125 X 0.125 & 45.13 & 0.079   & 0.130 & 0.266 & 0.475 & 2.377 \\ \hline
10  & 50                       & 250, 8bit & 0.125 X 0.125 & 15.10 & 0.199   & 0.326 & 0.266 & 0.790 & 1.581 \\ \hline
10  & 100                      & 250, 8bit & 0.125 X 0.125 & 5.10  & 0.397   & 0.651 & 0.266 & 1.315 & 1.342 \\ \hline
10  & 200                      & 250, 8bit & 0.125 X 0.125 & 0.15  & 0.805   & 1.306 & 0.268 & 2.380 & 1.190 \\ \hline
200 & 100                      & 82K, 4bit & 1.365 X 1.365 & 6.45  & 72.300  & 70.4  & 31.5  & 174.0 & 8.7   \\ \hline
200 & 50                       & 82K, 8bit & 1.925 X 1.925 & 14.05 & 79.500  & 70.8  & 64.0  & 214.0 & 21.4  \\ \hline
200 & 100 & 82K, 8bit & 1.925 X 1.925 & 4.07  & 153.000 & 141.0 & 63.8  & 358.0 & 17.9  \\ \hline
\end{tabular}
\label{tab:compperf}
\end{table*}

\begin{figure}
    \centering
    \includegraphics[width=\linewidth]{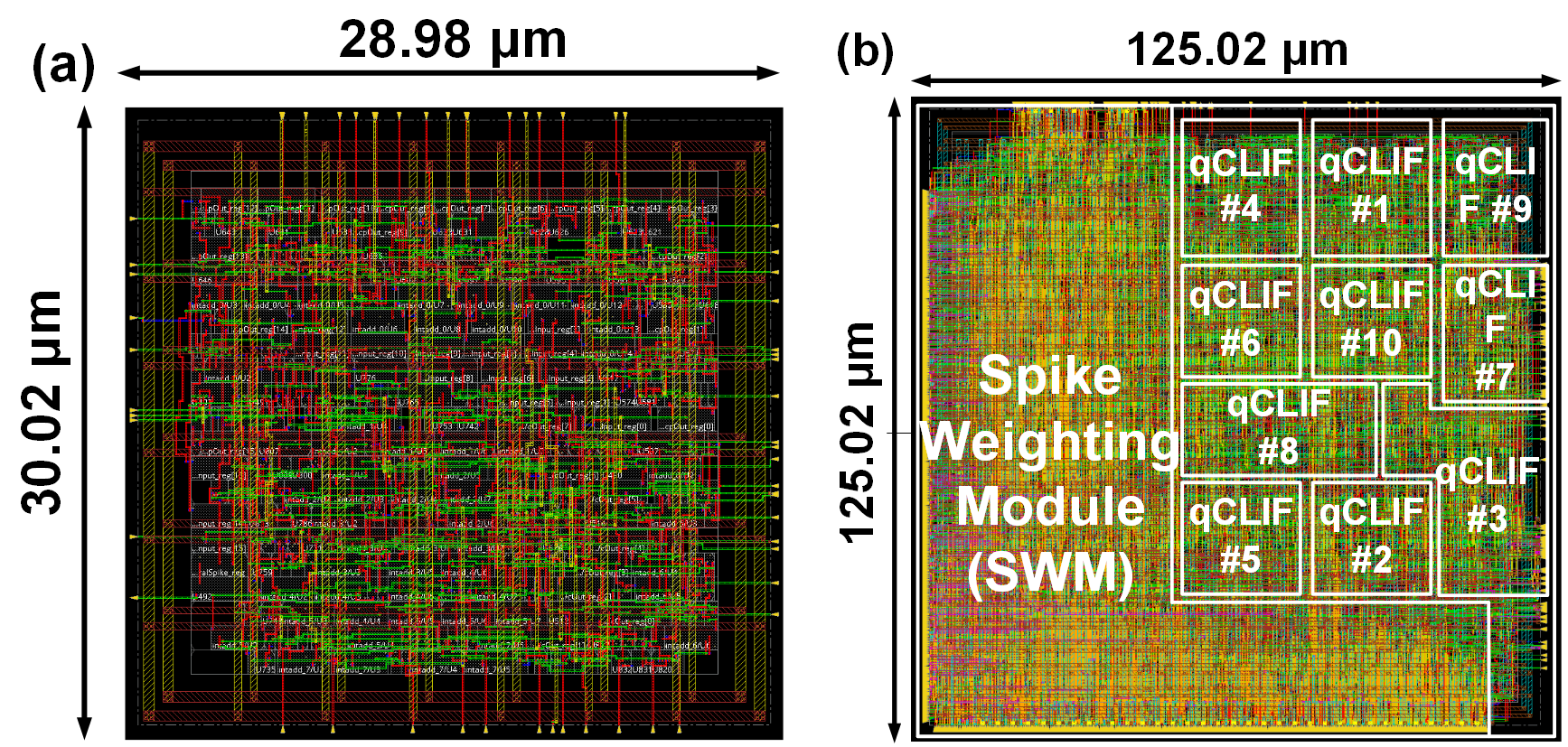}
    \caption{Layout of the proposed design: (a) single cLIF neuron, (b) a complete 10 neuron qCLIF model with weight accumulator.}
    \label{fig:layout}
\end{figure}

\begin{figure}
    \centering
    \includegraphics[width=\linewidth]{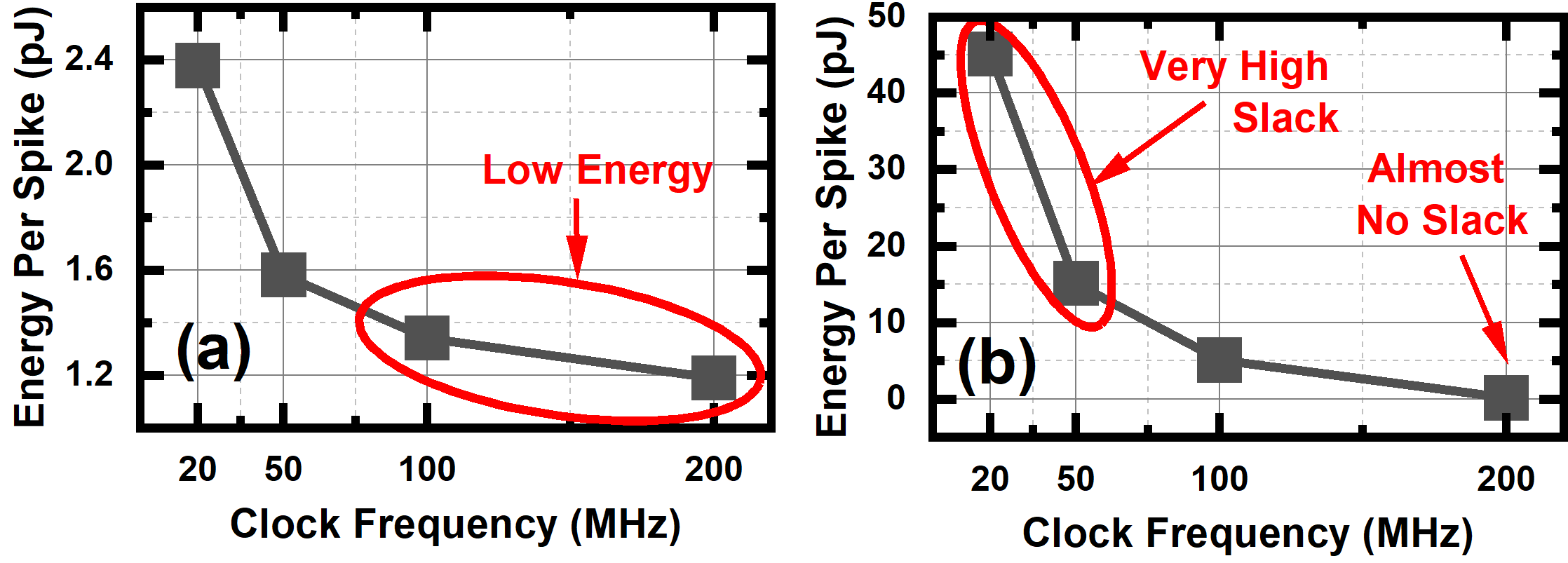}
    \caption{Performance evaluation of a 10 qCLIF neuron layer: (a) The relationship between clock frequency and energy per spike, showing lower energy efficiency at higher frequencies. (b) Clock frequency versus timing slack, illustrating the trade-off between speed and slack available for completing the operation.}
    \label{fig:fig8}
\end{figure}

The proposed qCLIF design, synthesized on a 45nm CMOS process \cite{10}, was realized through Synopsys Design Vision and Innovus automation tools. Simulation results for a single neuron with 8-bit precision indicated an area footprint of 0.029 × 0.030 mm² (layout in Fig. \ref{fig:layout} (a), a slack time of 5.62 ns, and power consumption metrics as follows: switching power at 0.020 mW, internal power at 0.041 mW, and leakage power at 0.016 mW. This yielded a total power consumption of 0.077 mW and an energy efficiency of 0.773 pJ per spike, as detailed in Table \ref{tab:compperf}. The performance metrics provided in this study primarily reflect worst-case scenarios and may not fully capture variations specific to different activities. It's plausible the networks exhibit sparse activity, hence could potentially demonstrate lower energy consumption. Fig. \ref{fig:layout} (b) shows a complete ten neuron qCLIF model layout with an accumulator. Further, all neurons are mapped at the right top part of the floor plan since a weight accumulator is located at the left part. Grouping the neurons in integrated circuit layouts enhances signal integrity by minimizing transmission distances and noise interference. This approach optimizes power distribution and reduces noise, while also streamlining the design and manufacturing process. This 10 qCLIF RSNN layer was subjected to simulations at various clock frequencies to evaluate network scalability. At a high clock frequency of 200 MHz, the network exhibited a minimal slack of 0.15 ns, albeit at the expense of increased power consumption. Conversely, a lower clock frequency of 20 MHz significantly reduced power consumption but increased the energy per spike. A median frequency of 100 MHz was found to offer a balanced trade-off, achieving a slack time of 5.10 ns and an energy per spike of 1.342 pJ as can be seen from \ref{fig:fig8}. 

To assess the scalability of the proposed design, it was applied to a larger network of 200 neurons, akin to a similar-sized RSNN used for DVS gesture classification, and operated at 100 MHz. With an 8-bit precision configuration, the network achieved a timing slack of 4.07 ns, occupying an area of 1.925 × 1.925 mm² and registering an energy consumption of 17.9 pJ per spike. A subsequent reduction in precision to 4 bits resulted in a smaller area of 1.365 × 1.365 mm² and lowered the energy per spike to 8.7 pJ. This demonstrates the design's adaptability to larger networks. Notably, the decrease in precision led to approximately a 50\% reduction in energy per spike and, interestingly, an increased timing slack, suggesting the potential for higher clock frequencies. As shown in Table \ref{tab:compperf}, both area and power requirements increase sub-linearly with the number of neurons and synapses, further indicating the scalability of the design methodology.

\begin{table*}[]
\centering
\caption{COMPARISON OF VARIOUS SNN HARDWARE}
\begin{tabular}{|c|c|c|c|c|c|c|c|}
\hline
 &
  \textbf{\cite{13}} &
  \textbf{\cite{14}} &
  \textbf{\cite{15}} &
  \textbf{\cite{16}} &
  \textbf{\cite{17}} &
  \textbf{\begin{tabular}[c]{@{}c@{}}This \\ work\end{tabular}} &
  \textbf{\begin{tabular}[c]{@{}c@{}}This \\ work\end{tabular}} \\ \hline
\textbf{\begin{tabular}[c]{@{}c@{}} \\ \end{tabular}}     & Fabricated     & Fabricated         & Fabricated    & Fabricated     & Fabricated     & \textbf{Simulated}    & \textbf{Simulated}    \\ \hline
\textbf{\begin{tabular}[c]{@{}c@{}}Technology \\ (nm)\end{tabular}}     & 65     & 90         & 65    & 10     & 28     & \textbf{45}    & \textbf{45}    \\ \hline
\textbf{\begin{tabular}[c]{@{}c@{}}Neuron \\ count\end{tabular}}        & 650    & 400        & 410   & 4096   & 1M     & \textbf{200}   & \textbf{200}   \\ \hline
\textbf{Network Type}                                                   & FF SNN & FF SNN     & SNN   & FF SNN & FF SNN & \textbf{cRSNN} & \textbf{cRSNN} \\ \hline
\textbf{Neuron Type}                                                    & IF     & Stochastic & IF    & LIF    & LIF    & \textbf{qCLIF} & \textbf{qCLIF} \\ \hline
\textbf{\begin{tabular}[c]{@{}c@{}}Synapse \\ count\end{tabular}}       & 67k    & 313k       & N//A  & 1M     & 256M   & \textbf{82k}   & \textbf{82 k}  \\ \hline
\textbf{Precision}                                                      & 6 bit  & 1bit       & 4 bit & 7 bit  & 4 bit  & \textbf{4 bit} & \textbf{8 bit} \\ \hline
\textbf{Area (mm2)}                                                     & 1.99   & 0.15       & 10.08 & 1.72   & 430    & \textbf{1.86}  & \textbf{3.71}  \\ \hline
\textbf{\begin{tabular}[c]{@{}c@{}}Clock \\ frequency\end{tabular}} &
  \begin{tabular}[c]{@{}c@{}}70KHz@\\    0.52V\end{tabular} &
  37.5MHz &
  20MHz &
  \begin{tabular}[c]{@{}c@{}}105MHz\\    @ 0.5V\end{tabular} &
  \begin{tabular}[c]{@{}c@{}}1KHz@\\    1.05V\end{tabular} &
  \textbf{\begin{tabular}[c]{@{}c@{}}100MHz@\\    1.1V\end{tabular}} &
  \textbf{\begin{tabular}[c]{@{}c@{}}100MHz@\\    1.1V\end{tabular}} \\ \hline
\textbf{\begin{tabular}[c]{@{}c@{}}Energy per \\ SOP (pJ)\end{tabular}} & 1.5    & 8.4        & N//A  & 3.8    & 26     & \textbf{8.7}   & \textbf{17.9}  \\ \hline
\textbf{Dataset} &
  \begin{tabular}[c]{@{}c@{}}GSCD \\ (4 Keywords)\end{tabular} &
  \begin{tabular}[c]{@{}c@{}}GSCD \\ (2 Keywords)\end{tabular} &
  \begin{tabular}[c]{@{}c@{}}GSCD \\ (10 Keywords )\end{tabular} &
  \begin{tabular}[c]{@{}c@{}}TIMIT \\ (4 Keywords)\end{tabular} &
  \begin{tabular}[c]{@{}c@{}}TDIGIT\\ (4 classes)\end{tabular} &
  \textbf{\begin{tabular}[c]{@{}c@{}}DVS Gesture\\ (10 Classes)\end{tabular}} &
  \textbf{\begin{tabular}[c]{@{}c@{}}DVS Gesture\\ (10 Classes)\end{tabular}} \\ \hline
\textbf{Accuracy (\%)}                                                  & 91.8   & 94.6       & 90.2  & 94     & 90.8   & \textbf{73}    & \textbf{90}    \\ \hline
\end{tabular}
\label{tab:compsnn}
\end{table*}

These results suggest that the qCLIF neuron model is a viable candidate for high-speed, energy-efficient neuromorphic computing applications. Compared to previous works, as seen in Table \ref{tab:compsnn}, while our design does not boast the highest neuron count, it introduces a more complex neuron model within a recurrent SNN framework for the first time while achieving lower energy per spike. Despite the effectiveness of this approach, it deviates from the asynchronous operation ideal in fully digital neuromorphic systems, as synchronized functioning is required for both apical and somatic compartments. The design's reliance on a digital accumulator, occupying substantial layout space, suggests potential for refinement. Space-efficient alternatives, such as sparse accumulator or memristor crossbar architectures output can be directly connected to the qCLIF digital hardware. The exploration of smaller technology nodes could yield further improvements. 

\section{conclusion}

In this study, we propose a qCLIF neuron model featuring variable precision utilizing networks sparse activity. We implemented a scalable, reconfigurable qCLIF neuron layer, marking the first hardware realization of a context-based recurrent spiking neuron layer in the digital domain. These designs are evaluated at a 45nm technology node through synthesis and layout. Our evaluations across different operating frequencies aimed to balance computational efficiency and hardware performance optimally. This work lays a step towards digital efficient neuromorphic hardware systems.

\section*{Acknowledgment}

The authors appreciate discussions with R. Legenstein, G. H. Hutchinson, T. Bhattacharya. This study is supported by the USA National Science Foundation award \#2318152.

\end{document}